\documentclass{article}
\usepackage{spconf,amsmath,graphicx}
\usepackage{setspace}
\usepackage{bm}

\usepackage{comment}

\usepackage {booktabs}
\usepackage {colortbl,array,xcolor}

\usepackage{graphicx} 
\usepackage{booktabs} 
\usepackage{multirow} 
\usepackage{xcolor} 
\usepackage{amsmath} 
\usepackage{tabularx}

\usepackage{bbding}
\usepackage{pifont}
\usepackage{wasysym}
\usepackage{amssymb}

\usepackage{cite}

\title{Triplet Synthesis For Enhancing Composed Image Retrieval via Counterfactual Image Generation}
\name{Kenta Uesugi, Naoki Saito, Keisuke Maeda, Takahiro Ogawa, Miki Haseyama}
\address
    {Hokkaido University, Japan\\
    E-mail:\{uesugi, saito, maeda, ogawa, mhaseyama\}@lmd.ist.hokudai.ac.jp}

\begin{document}
\ninept

\maketitle

%
\begin{abstract}
Composed Image Retrieval (CIR) provides an effective way to manage and access \mbox{large-scale} visual data. 
Construction of the CIR model utilizes triplets that consist of a reference image, modification text describing desired changes, and a target image that reflects these changes.
For effectively training CIR models, extensive manual annotation to construct \mbox{high-quality} training datasets, which can be \mbox{time-consuming} and \mbox{labor-intensive}, is required. 
To deal with this problem, this paper proposes a novel triplet synthesis method
by leveraging counterfactual image generation. 
By controlling visual feature modifications via counterfactual image generation, our approach automatically generates diverse training triplets without any manual intervention.
This approach facilitates the creation of larger and more expressive datasets, leading to the improvement of CIR model's performance. 
\end{abstract}


\renewcommand{\thefootnote}{\fnsymbol{footnote}}
\footnote[0]{This work was partly supported by JSPS KAKENHI Grant Numbers JP24K02942, JP23K21676 and JP23K11211.}
\renewcommand{\thefootnote}{\arabic{footnote}}

\vspace{-3mm}
\section{Introduction}
\label{sec:intro}
The explosion of digital content has made the efficient access to and management of vast amounts of visual information increasingly essential. 
As the volume of visual information grows, users are seeking more advanced tools to find their desired information. 
Recently, there has been a growing demand for image retrieval that can interpret and identify desired images based on human visual instructions.

In response to the growing demand, various image retrieval methods have been proposed in previous studies~\cite{el2021training,song2023boosting,saito_2023_pic2word,Baldrati_2023_ICCV,gu2024compodiff,gu2024lincir,baldrati2023composed,zhang2024zero}.
Traditional image retrieval methods~\cite{el2021training,song2023boosting} solely rely on images as input, which limits their effectiveness in handling complex queries. 
Specifically, these methods often struggle to distinguish visually similar items with distinct attributes. 
They also have difficulty adapting to specific modifications.
Since instructions with images are difficult to align with user's intentions, the utility is restricted in scenarios that need more precise and flexible retrievals.
To address these limitations, Composed Image Retrieval (CIR)~\cite{Baldrati_2023_ICCV,zhang2024zero,baldrati2023composed,saito_2023_pic2word, gu2024lincir,gu2024compodiff} has been proposed. 
CIR allows users to start with a reference image and then refine the retrieval results by specifying modifications or additional details in textual form. 
Since this \mbox{dual-input} approach enhances retrieval with flexibility, it is especially valuable in industries such as \mbox{e-commerce}, where users often seek products with specific characteristics~\cite{chen2024spirit,chen2023real20m,kuang2019fashion}.

While CIR shows great potential for improving image retrieval experiences, there are still significant challenges in applying it to \mbox{real-world} applications. 
One of the most critical issues is the difficulty in collecting training data to construct CIR models.
Specifically, the construction of them requires \mbox{large-scale} datasets composed of the following three components: a reference image, modification text, and a target image, known as triplets. 
The collection of these triplets is typically costly and traditionally relies on manual annotation~\cite{Wu_2021_CVPR,liu2021image}, which makes it difficult to gather the \mbox{large-scale} datasets necessary for practical CIR model training.
To deal with this issue, Ventura et al. have proposed an automatic method to select image pairs for triplets from captions previously assigned to the \mbox{large-scale} image dataset~\cite{ventura23covr}. 
However, this automatic triplet collection method has several critical issues. 
This method focuses solely on collecting similar images based on their captions, which may obtain \mbox{low-quality} triplets.
That is, the pairs of images for triplets differ significantly in aspects not described by the modification text.
In the training data for CIR, \mbox{high-quality} triplets that accurately reflect only the modified parts are required.
Furthermore, when the image set is limited, finding appropriate image pairs becomes much more difficult.
Namely, it is difficult to collect image pairs that reflect only the desired modifications, leading to inadequate triplet construction.
Consequently, it is necessary to develop a method that can construct \mbox{high-quality} triplets in \mbox{data-scarce} scenarios.

In this paper, we present a new triplet synthesis method for enhancing CIR performance. 
Our method focuses on counterfactual image generation models~\cite{prabhu2023lance,li2024reinforcing,le2024coco} as a solution to the challenges in triplet construction. 
Counterfactual image generation models incorporate the concept of counterfactual reasoning, which involves assuming a situation that did not actually occur to gain a deeper understanding of the current one. 
These models make slight edits to the input image to reflect a hypothetical scenario and generate a counterfactual image.
They can generate a new image with different attributes while maintaining certain relationships between this image and the input image. 
By using counterfactual image generation, it is possible to synthesize triplets that were not initially present in the dataset, thereby enhancing the diversity of the dataset.
Furthermore, although the generated images are semantically different from the original ones, they still retain some relevance, which helps the training CIR models better learn the difference between the two images.
In addition, by focusing on the visual differences caused by local changes, the retrieval model becomes more sensitive to the details of the subject. 
As a result, it can detect minute semantic changes that cannot be captured by conventional methods and is expected to improve retrieval performance.

\begin{figure*}[!t]
    \centering
    \includegraphics[width=0.995\linewidth]{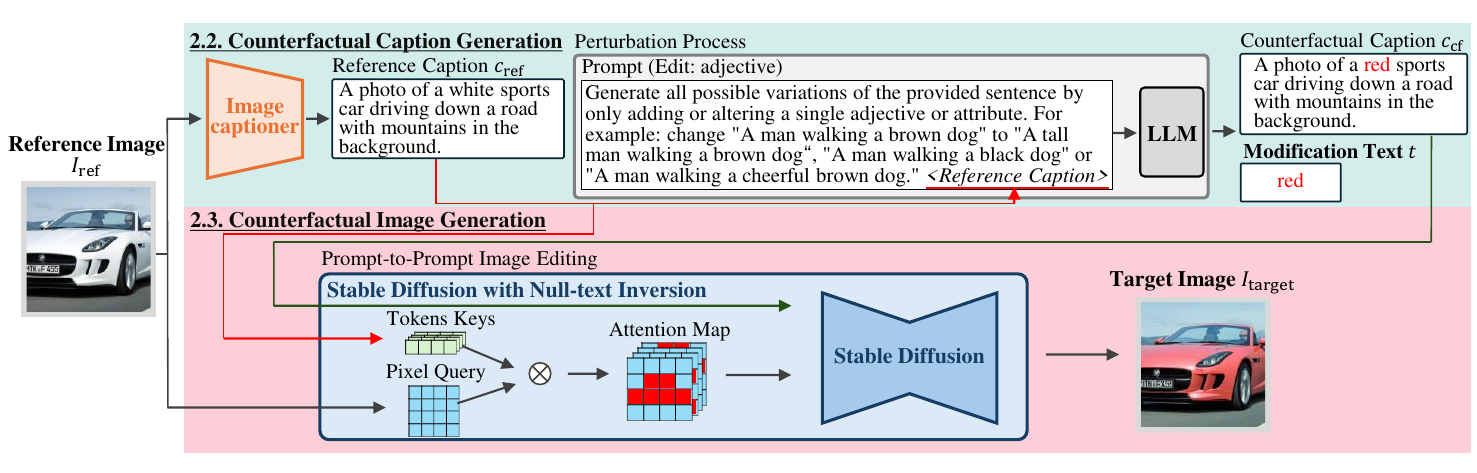}
    \vspace{-4mm}
    \setlength{\abovecaptionskip}{0mm} 
    \setlength{\belowcaptionskip}{0mm} 
    \vspace{-3mm}
    \caption{Overview of the proposed triplet synthesis method for enhancing CIR performance.
    This figure shows the process of synthesizing triplets. 
    It involves modifying the reference caption $c_{\rm{ref}}$ to create a counterfactual caption $c_{\rm{cf}}$ and generating its corresponding image. 
    \label{fig:overview}}
    \vspace{-4mm}
\end{figure*}

In summary, the key contribution of this paper is the introduction of the counterfactual image generation.
The proposed method enables the synthesis of \mbox{high-quality} triplets based on slight modifications without any manual annotation. 
Even with limited image pools, our method preserves both the quality and quantity of reference and target images, and it generates diverse triplets from the restricted data pool.
Consequently, the proposed method can overcome traditional challenges in automatic triplet synthesis for CIR and facilitate the generation of practical and \mbox{large-scale} CIR datasets.
The main contributions of this paper are shown as follows.

\begin{itemize}
  \item  We introduce a novel triplet synthesis method for CIR using the counterfactual image generation, which enhances the \mbox{high-quality} and diversity of triplets.

  \item We demonstrate that this method can significantly improve CIR model performance even when only a \mbox{small-scale} dataset is available.
\end{itemize}


\section{Triplet synthesis using counterfactual image generation}

\label{sec:method}
\subsection{Overview of our method}
The overview of our triplet synthesis method is presented in Fig.~\ref{fig:overview}. 
The objective of this task is to acquire a triplet $\langle I_{\rm{ref}}, t, I_{\rm{target}} \rangle$, which serves as the training data for CIR.
Here, $I_{\rm{ref}}$ represents a reference image, $t$ denotes modification text, and $I_{\rm{target}}$ is a target image.
The target image $I_{\rm{target}}$ is generated by applying the modification text $t$ to the reference image $I_{\rm{ref}}$.

The proposed method utilizes \mbox{Language-guided} Counterfactual Image (LANCE)~\cite{prabhu2023lance} to generate the counterfactual image as the target image $I_{\rm{target}}$.
While LANCE was originally designed to generate images to evaluate classification models, we have extended this model to support the triplet synthesis.
The triplet synthesis using LANCE involves two primary steps: counterfactual caption generation and its image generation.
%
First, LANCE generates a counterfactual caption $c_{\rm{cf}}$ describing hypothetical situations for the reference image $I_{\rm{ref}}$.
Then, based on the counterfactual caption and reference image, a counterfactual image that serves as the target image $I_{\rm{target}}$ is generated.
The target image $I_{\rm{target}}$ undergoes minimal modifications while maintaining its relevance to the reference image $I_{\rm{ref}}$, and these modifications correspond to the modification text $t$.

Even if the number of training images is limited, our method enables the synthesis of the necessary triplets for the model training.
Generating image pairs with small changes helps the CIR model better understand the relationships between the images.
Through the comprehensive process illustrated in Fig.~\ref{fig:overview}, our method synthesizes triplets that could not have existed in the original dataset and allows for the construction of more diverse and \mbox{high-quality} triplets for training the CIR model.

\subsection{Counterfactual Caption Generation}
This stage requires the sufficiently meaningful caption $c_{\rm{ref}}$ to explain the reference image $I_{\rm{ref}}$.
Thus, to generate the reference caption $c_{\rm{ref}}$, the proposed method utilizes \mbox{BLIP-2~\cite{li2023blip}}, a model to provide rich and descriptive captions that capture essential details in the same manner as LANCE's caption generation process. 

After generating the reference caption $c_{\rm{ref}}$, our method applies targeted perturbations to specific components of the reference caption $c_{\rm{ref}}$. 
These elements include crucial parts that define the visual content of the reference image $I_{\rm{ref}}$, such as the subject, object, background, adjectives, and domain.
For example, suppose the reference caption $c_{\rm{ref}}$ describes ``a photo of a white sports car driving down a road with mountains in the background.''
In this case, the perturbation might change ``white'' to ``red'' or ``mountains'' to ``buildings.''
This perturbation process uses a \mbox{fine-tuned} LLM that is designed to change only a part of the input sentence. 
When the reference caption $c_{\rm{ref}}$ is input into this \mbox{fine-tuned} LLM, it outputs the modification text $t$ along with the counterfactual caption $c_{\rm{cf}}$ after the perturbation has been applied.
Importantly, these modifications adjust only a part of the caption, which results in changes to only that part of the image.
This level of control is crucial for maintaining the integrity of the triplets used for training CIR models.

\subsection{Counterfactual Image Generation}
The counterfactual caption $c_{\rm{cf}}$ obtained in the previous subsection is used as input to a \mbox{text-to-image} generation model, \textit{i.e.}, Stable Diffusion~\cite{rombach2022high}, to generate the counterfactual image. 
Stable Diffusion is a latent diffusion model known for generating high-quality images based on text input.
However, even minor modifications in the prompt can cause significant, often undesirable, alterations in the generated image.
To deal with this issue, our method leverages the \mbox{prompt-to-prompt} image editing technique~\cite{hertz2022prompt}, which enables targeted adjustments by injecting \mbox{cross-attention} maps corresponding to the caption edits at certain stages of the diffusion process. 
Additionally, since applying this technique to authentic images requires accurate inversion to the latent space, we utilize the \mbox{null-text} inversion method~\cite{mokady2023null}, which enables precise reconstruction of the original image in the latent space. 
This process reverses the diffusion trajectory while maintaining close alignment with the original image encoding. 
By employing these techniques, our method ensures that the counterfactual image remains faithful to the original visual content, accurately reflecting the intended textual perturbation without introducing unwanted artifacts or distortions.

There is an advantage to using counterfactual image generation models for triplet synthesis in CIR.
Specifically, triplets based on authentic images often contain a lot of unnecessary information beyond the intended changes.
This excess information can make it difficult for the CIR model to learn the necessary information effectively.
Additionally, there is a higher risk that the model might pick up on unintended patterns or correlations from these extra features.
In contrast, by using triplets made up of image pairs generated from counterfactual image generation models, the CIR model can focus on learning the specific changes intended by the user. 
This allows the CIR model to filter out unnecessary changes and focus on the intended modifications, which is expected to improve the performance of the CIR model.



\vskip\baselineskip


\section{Experimental Results}
\label{sec:experiment}

\subsection{Settings}
\vspace{0mm}
To verify the effectiveness of our method, we utilized two datasets: Composed Image Retrieval with Reasoning (CIRR) dataset~\cite{liu2021image}, which includes complex query images in natural scenes, and FashionIQ~\cite{Wu_2021_CVPR} dataset, which focuses on fashion items.
Each dataset is intended to verify the ability to handle complex queries.
%
%
CIRR dataset, derived from the NLVR2~\cite{suhr2018corpus}, challenges models to reason about complex visual relationships. 
FashionIQ dataset composes images of various clothing items, \textit{i.e.}, shirts, dresses, and tops/tees. 
We synthesized \mbox{5,000 triplets} from CIRR dataset, and \mbox{3,000 triplets} from FashionIQ dataset using \mbox{1,500 training} images from each dataset, respectively.

\begin{table}[t!] 
    \caption{\small Summary of dataset statistics.
    The training data of CIRR and FashionIQ datasets were reduced to 30\% of their original number of images to simulate data-scarce scenarios. 
    }
    \vspace{2mm}
    \centering
    \footnotesize
    {
    \begin{tabular}{lcccc}
        \toprule
        \multirow{2}{*}{Dataset}& \multicolumn{2}{c}{Images}  & \multicolumn{2}{c}{Triplets}\\
        \cmidrule(lr){2-3} \cmidrule(lr){4-5}
         &Train &Test &Train / Synthetic Data &Test \\ 
        \midrule
        CIRR &5,082 &2,265 &1,392 / 5,000 &4,148 \\ 
        FashionIQ &13,623  &15,415 &1,487 / 3,000 &6,016  \\
        \bottomrule
    \end{tabular}
    }
    \label{tab:dataset}
    \vspace{-4mm}
\end{table}


\begin{table*}[ht!]
    \caption{Recall (\%) of retrieval results on CIRR and FashionIQ datasets.
    When the retrieval performance improves with our method in each retrieval model, the result of R@\textit{k} is shown in \textbf{bold}.
    }
    \vspace{2mm}
    \centering
    \footnotesize 
    \begin{tabular}{llcccccccc}
        \toprule[1.0pt]
        \multirow{2}{*}{} & \multirow{2}{*}{Retrieval method}  & \multicolumn{4}{c}{CIRR} & \multicolumn{2}{c}{FashionIQ} \\
        \cmidrule(lr){3-6} \cmidrule(lr){7-8} 
        & & R@1 & R@5 & R@10 & R@50 & R@10 & R@50 \\
        \midrule[0.05em]
        \multirow{4}{*}{Zero-shot} & Pic2Word (Saito '23)~\cite{saito_2023_pic2word} & 23.90 & 51.70 & 65.30 & 87.80 & 24.70 & 43.70 \\
                                   & SEARLE (Baldrati '23)~\cite{Baldrati_2023_ICCV} &24.24 &52.48 &66.29 &88.84 &25.56 &46.23 \\
                                   & CompoDiff (Gu '23)~\cite{gu2024compodiff} &19.37 &53.81 &72.02 &90.85 &37.36 &50.85 \\
                                   & LinCIR (Gu '24)~\cite{gu2024lincir}  &30.89 &60.75  &73.88 &92.84 &26.39  &46.71  \\
        \midrule[0.05em]
        \multirow{2}{*}{Baseline (w/o synthetic triplets)} 
                               & Combiner (Baldrati '23)~\cite{baldrati2023composed} &32.65 &63.78 &75.59 &93.28 &31.93 &54.20 \\
                               & BLIP (Ventura '24)~\cite{ventura23covr} &39.66 &68.07 &78.17 &93.01 &36.34 &58.27 \\
        \midrule[0.05em]
        \multirow{2}{*}{Our method} 
                               & Combiner (Baldrati '23)~\cite{baldrati2023composed} &32.58 &63.78 &\textbf{75.60} &\textbf{94.12}&31.92&\textbf{54.40} &\\
                               & BLIP (Ventura '24)~\cite{ventura23covr} &\textbf{40.75} &\textbf{69.83} &\textbf{81.04} &\textbf{94.80} &\textbf{39.13} &\textbf{60.61} \\
        \bottomrule[1.0pt]
    \end{tabular}
    \label{tab:results_1}
\end{table*}

This experiment evaluated the performance of our method through a series of image retrieval tasks.
To evaluate image retrieval performance, we used Recall@\textit{k} (R@\textit{k}), a standard metric for retrieval tasks that measures the percentage of relevant items found in the \mbox{top-\textit{k}} retrieved results. 
In addition, this experiment used \mbox{LLAMA-7B}~\cite{touvron2023llama} with LoRA \mbox{fine-tuning}~\cite{hu2021lora} for structured caption perturbations. 
In this experiment, we employed two kinds of evaluations: comparisons of different retrieval methods under the condition of a small amount of data in subsection~3.2, 
and an ablation study by varying the number of original training images in subsection~3.3.

\begin{figure*}[!t]
    \centering
    \includegraphics[width=0.95\linewidth]{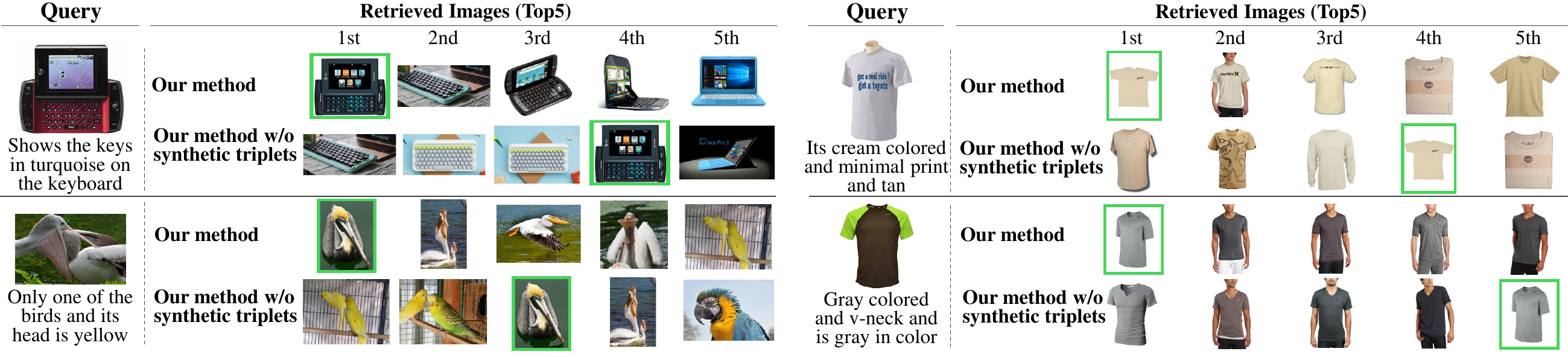}
    \vspace{-2mm}
    \setlength{\abovecaptionskip}{0mm} 
    \setlength{\belowcaptionskip}{0mm} 
    \vspace{-1mm}
    \caption{Retrieved results on the CIRR and FashionIQ datasets.
    It highlights that the proposed synthetic triplet method enables the acquisition of essential characteristics of images.
    Note that the ground truth image is indicated by a green box.
    \label{fig:retrieval_result}}
    \vspace{-3mm}
\end{figure*}

\begin{figure*}[!h]
    \centering
    \scalebox{1.0}{
    \includegraphics[width=0.85\linewidth]{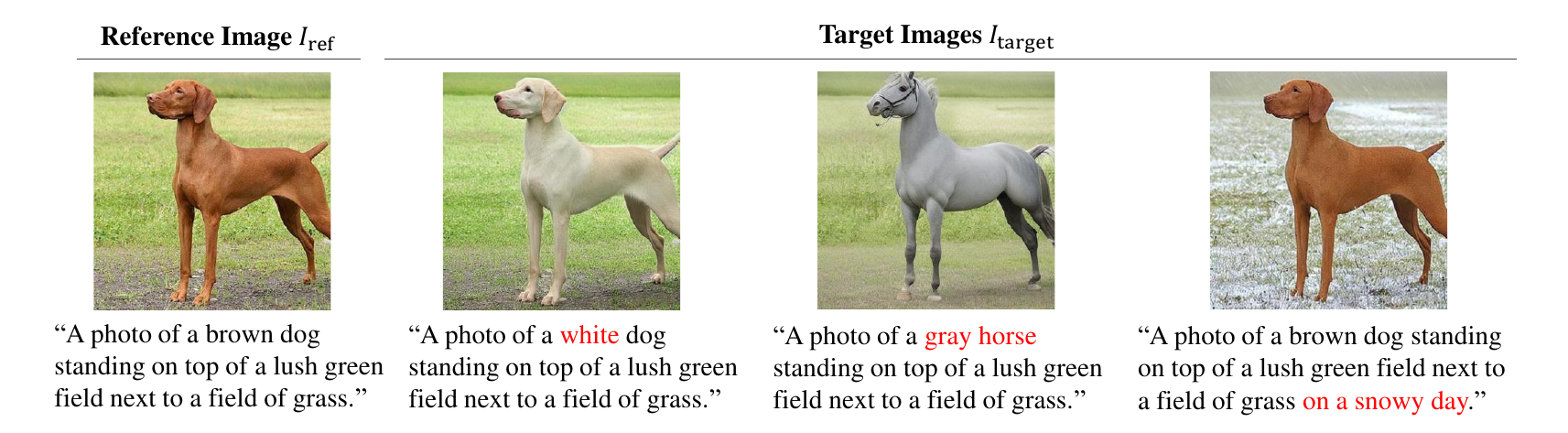}
    }
    \vspace{-3mm}
    \setlength{\abovecaptionskip}{0mm} 
    \setlength{\belowcaptionskip}{0mm} 
    \vspace{-2mm}
    \caption{Examples of triplets synthesized by the proposed method.
    This figure shows an example of target image generation modified by the modification text shown in red.
    The global structure remains intact, while only local changes are reflected in the synthesized images.
    \label{fig:generate_examples}}
    \vspace{-3mm}
\end{figure*}

\subsection{Evaluation of Composed Image Retrieval Performance}

This subsection verifies the effectiveness of the proposed method by comparing the image retrieval performance of the previous CIR methods. 
The number of images and triplets of the CIRR and FashionIQ datasets using this evaluation is shown in Table~\ref{tab:dataset}. 
This evaluation performed CIR with a limited dataset, and we conducted this experiment with a reduced dataset, where the number of images was decreased to 30\% of the original training dataset. 
Then, the triplets were constructed from these limited images. 
Note that CIRR dataset used test data, and FashionIQ dataset used validation data to obtain retrieval results.
This choice, along with the values of R@$k$ for each dataset, was made to maintain consistency with the evaluation methods used in prior research~\cite{ventura23covr}.
In this evaluation, we used \mbox{fine-tuned} BLIP model~\cite{ventura23covr} and Combiner~\cite{baldrati2023composed} as CIR models. 
In addition, to compare the performance improvement across different CIR methods, this evaluation utilized several comparative methods. 
Specifically, as zero-shot CIR methods, we utilized Pic2Word~\cite{saito_2023_pic2word}, SEARLE~\cite{Baldrati_2023_ICCV}, CompoDiff~\cite{gu2024compodiff}, and LinCIR~\cite{gu2024lincir}. 
Furthermore, this evaluation used the proposed method without synthetic triplets to verify the effectiveness of our method. 

Table~\ref{tab:results_1} presents the CIR results on the CIRR and FashionIQ datasets.  
These experimental results show that the use of synthetic triplets consistently enhances the performance of the CIR models. 
Our results indicate that the proposed method significantly improves retrieval accuracy in \mbox{data-scarce} scenarios, highlighting its effectiveness in enhancing model performance when original data is limited.

The qualitative retrieval results are shown in Fig.~\ref{fig:retrieval_result}. 
It is evident that the model without counterfactual image generation (bottom) struggles to accurately capture the characteristics of the subject. 
On the other hand, the proposed method, which incorporates \mbox{high-quality} triplets with localized changes into the training process, can capture object characteristics more precisely. 
As a result, the retrieval performance is improved, and these results are more appropriate for the user's input query.

Figure~\ref{fig:generate_examples} illustrates examples of synthetic triplets by our method with the CIRR dataset. 
These examples show how the method accurately modifies reference images based on modification text to produce target images that reflect specified changes.
In the first example, the substitution of ``brown'' with ``white'' demonstrates the model's capability to target and alter specific objects without affecting the surrounding environment. 
This is essential for CIR, where users often seek to refine retrieval based on detailed modification text.
The examples also demonstrate that the synthetic triplets maintain consistency in unaltered elements, such as the background, while introducing localized changes. 
This consistency helps the model distinguish between subtle and significant changes, enhancing its ability to retrieve images that closely match complex and \mbox{user-specified} criteria.
Thus, the triplets synthesized by our method are ideal for training CIR models, as they provide diverse yet controlled examples that challenge the model to understand and apply nuanced visual distinctions.



\begin{figure}[!t]
    \centering
    \includegraphics[width=1.0\linewidth]{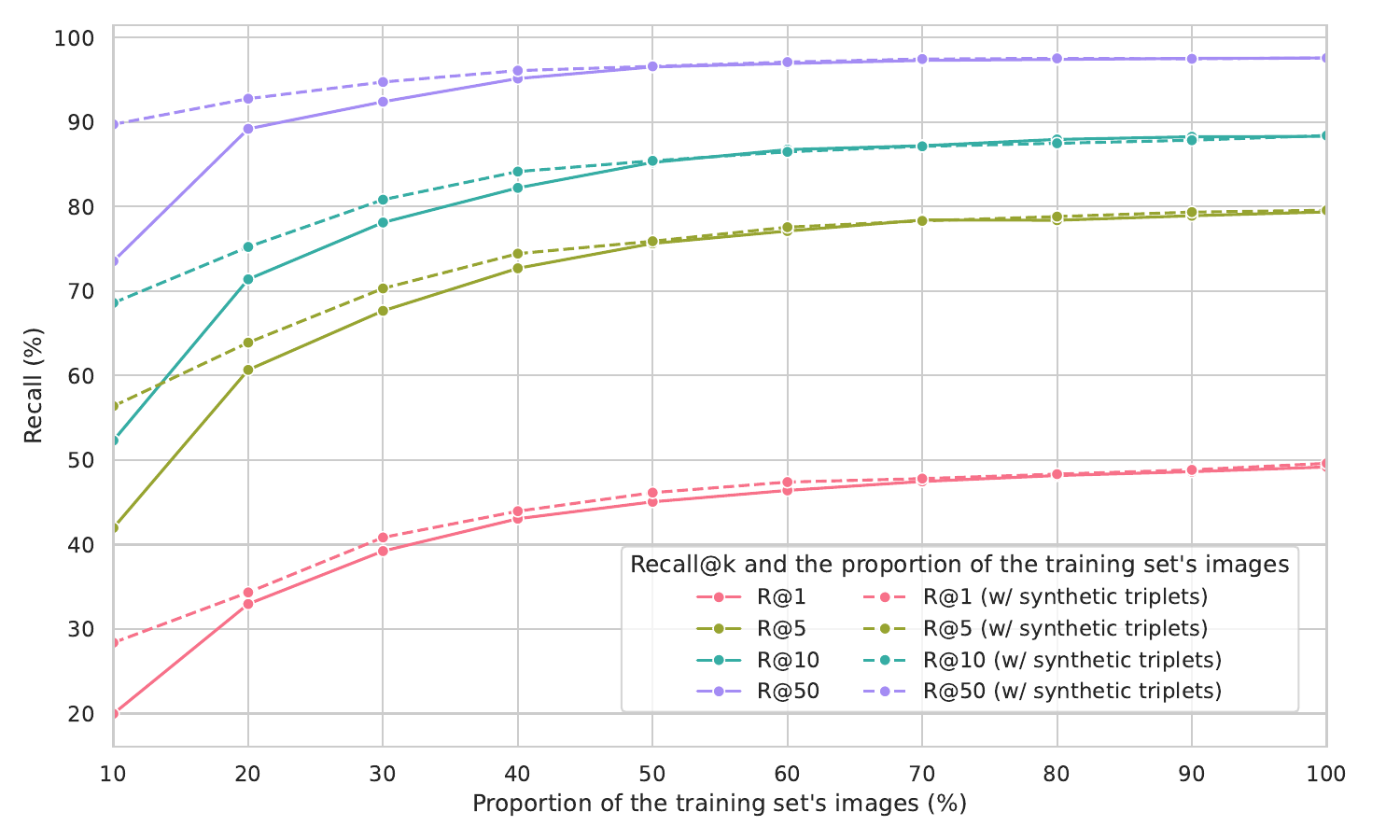}
    \vspace{-6mm}
    \setlength{\abovecaptionskip}{0mm} 
    \setlength{\belowcaptionskip}{0mm} 
    \vspace{-3mm}
        \caption{Comparison of R@\textit{k} across different training image proportions and synthetic triplets.
        This figure shows R@1, R@5, R@10, and R@50 for different numbers of training images, with and without the addition of 5,000 synthetic triplets.\label{fig:data_ablation}}
    \vspace{-5mm}
\end{figure}


\subsection{Ablation Study with Variable Training Data}
In this subsection, we conducted an ablation study by modifying the reduction proportion of the number of images in the original training dataset, which was fixed at 30\% in the previous experiment, as shown in Table~\ref{tab:dataset}.
In this evaluation, we demonstrated how our triplet synthesis method improves CIR performance.
This evaluation was conducted using the test set of the CIRR dataset.
We utilized \mbox{fine-tuned} BLIP model as the CIR model proposed in the previous research~\cite{ventura23covr}. 
Specifically, this evaluation was conducted by fixing the number of synthetic triplets at 5,000 and varying the ratio of the number of original images of training set in the CIRR dataset. 
Note that the number of original images is 16,939, and the number of manually constructed triplets from all of these images is 28,225 in this evaluation.
We evaluated the effectiveness of these synthetic triplets for CIR by comparing the image retrieval results with and without the inclusion of these \mbox{5,000 synthetic} triplets.

As depicted in Fig.~\ref{fig:data_ablation}, these results strongly reflect the effectiveness of the proposed method. 
Specifically, the addition of \mbox{5,000 synthetic} triplets consistently boosted the recall at any different levels~($k\in\{1, 5, 10, 50\}$)~compared to using the original training triplets alone. 
This effect was particularly noticeable in scenarios with smaller training images, where triplet synthesis significantly improved retrieval performance.
%
These results indicate that even with limited data, triplet synthesis can help the model capture essential attributes that might be otherwise overlooked.


\section{Conclusions}
\label{sec:conclusion}
In this paper, we have proposed a method to improve the performance of CIR through automatic triplet synthesis using counterfactual image generation.
Our method enhances the ability to generate diverse and \mbox{high-quality} training examples, advancing the capabilities of the CIR model.
Experimental results using CIRR and FashionIQ datasets confirmed that our method effectively created the diverse and \mbox{high-quality} triplets, including the counterfactual images that accurately reflect intended modifications. 


\vfill\pagebreak

\clearpage
\setstretch{0.99}
\bibliographystyle{IEEEbib}

\end{document}